\documentclass[runningheads]{llncs}
\usepackage{graphicx}
\usepackage{hyperref}

\begin{document}

\title{Synthetic Data Generation for 3D Myocardium Deformation Analysis}
\author{Shahar Zuler \and Dan Raviv}
\authorrunning{S. Zuler and D. Raviv}
\institute{Tel Aviv University, Tel Aviv, Israel \\ \email{shahar.zuler@gmail.com, darav@tauex.tau.ac.il}}
\maketitle

\begin{abstract}
Accurate analysis of 3D myocardium deformation using high-resolution computerized tomography (CT) datasets with ground truth (GT) annotations is crucial for advancing cardiovascular imaging research. 
However, the scarcity of such datasets poses a significant challenge for developing robust myocardium deformation analysis models. 
To address this, we propose a novel approach to synthetic data generation for enriching cardiovascular imaging datasets. 

We introduce a synthetic data generation method, enriched with crucial GT 3D optical flow annotations.
We outline the data preparation from a cardiac four dimensional (4D) CT scan, selection of parameters, and the subsequent creation of synthetic data from the same or other sources of 3D cardiac CT data for training.

Our work contributes to overcoming the limitations imposed by the scarcity of high-resolution CT datasets with precise annotations, thereby facilitating the development of accurate and reliable myocardium deformation analysis algorithms for clinical applications and diagnostics.

Our code is available at: \url{www.github.com/shaharzuler/cardio_volume_skewer}


\end{abstract}

\begin{figure}[h] 
    \centering
        \includegraphics[width=0.5\linewidth]{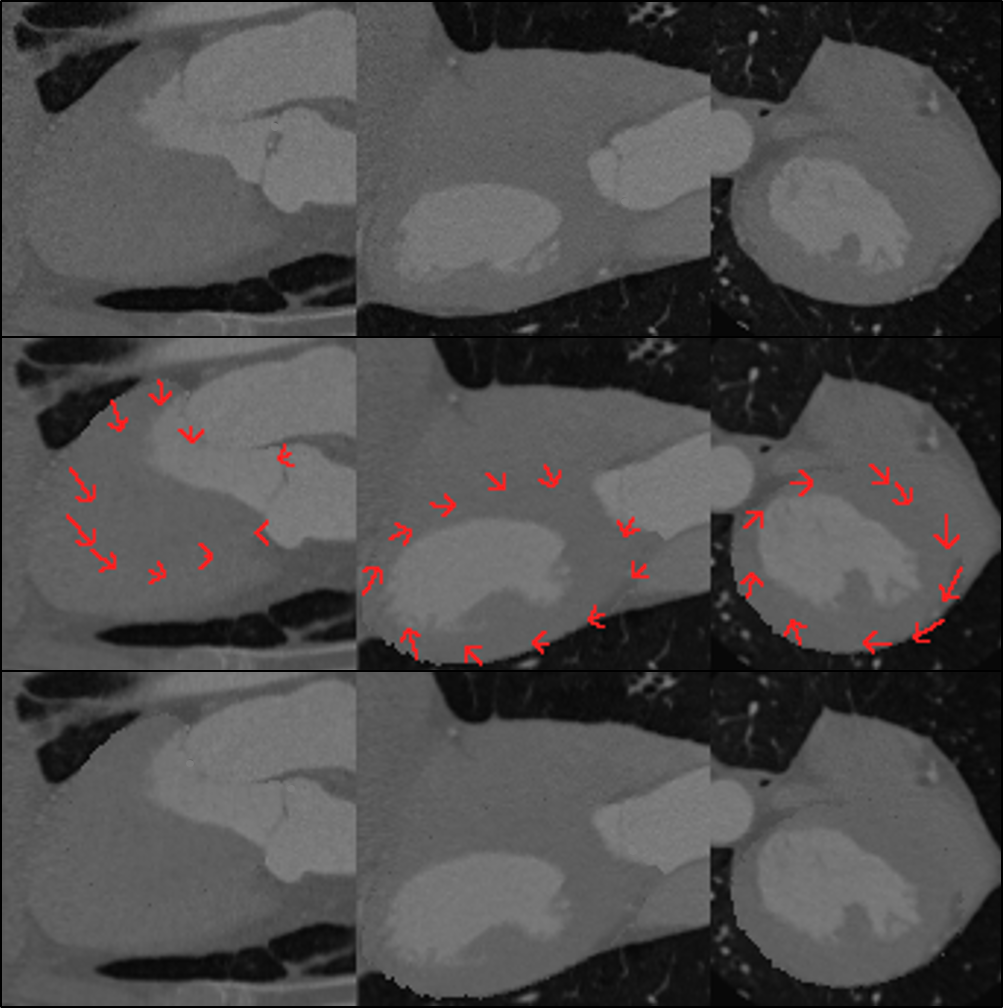}
        \caption{Example of Synthetic Sample: Visualization of a synthetic sample before (top) and after (bottom) deformation, along with arrows representing selected ground truth annotations (middle).}
    \label{fig:sample}
\end{figure}

\section{Introduction}

The accurate analysis of 3D myocardium deformation using CT datasets with GT annotations is integral to advancing cardiovascular imaging research. However, the scarcity of high-resolution CT datasets featuring precise annotations poses a significant challenge for developing robust myocardium deformation analysis models. A thorough understanding of myocardial changes throughout the cardiac cycle is essential for various clinical applications and diagnostics.

In this context, the motivation for our work becomes evident. We aim to address the pressing need for comprehensive datasets by introducing a novel approach to synthetic data generation. The synthesis of datasets with crucial GT annotations is essential for overcoming the limitations imposed by the current scarcity of such datasets. We delve into the challenges associated with obtaining high-quality datasets, paving the way for more accurate and reliable myocardium deformation analysis algorithms.

Given the challenges in obtaining CT datasets with accurate 3D optical flow annotations, we introduce our innovative synthetic data generation process. This approach enriches datasets and plays a significant role in developing precise models for analyzing myocardium deformation.

\section{Related Work}

In this section we describe existing methods for creation of datasets containing 3D deformation GT annotations.
 \subsection{Speckle Tracking}
 
 This method, commonly employed in ultrasound (US) imaging, leverages the interaction of ultrasound waves with myocardial tissue to create distinctive speckle patterns that can be tracked over time. These speckle patterns serve as unique features within the heart and have been employed for assessing various aspects of cardiac mechanics, including the analysis of cardiac strain and its clinical applications ~\cite{abduch2014cardiac}. However, the reliability of speckle tracking analysis is influenced by operator expertise and is sensitive to noise. Additionally, it may face limitations in assessing deeper cardiac structures.

\subsection{Magnetic Resonance (MR) Tagging}

MR tagging involves the introduction of a grid or pattern of dark lines during image acquisition, providing a dynamic reference system for evaluating myocardial deformation and strain. While effective, MR tagging can be time-consuming and requires expertise, as well as access to suitable Magnetic Resonance Imaging (MRI) equipment. Moreover, limited through-plane resolution in MRI scans may impact accuracy when compared to CT scans. Further insights into this method and its application in evaluating pathologies can be found in ~\cite{jeung2012myocardial}.

\section{Coordinate System}

\begin{figure}[h] 
    \centering
        \includegraphics[width=0.7\linewidth]{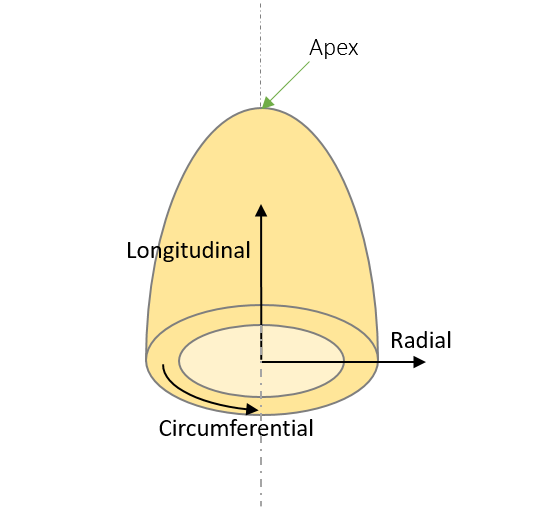}
        \caption{Visualization of coordinate system. The longitudinal direction is established as the principal component of the LV, intersecting the center of mass of the LV. Radial direction lines are perpendicular to the principal component, while circumferential direction lines are normal to both the intersecting radial line and the principal component itself.}
    \label{fig:coordinates}
\end{figure}

The myocardium exhibits diverse movements within the LV, encompassing longitudinal, radial, and circumferential motions, as visually represented in Figure \ref{fig:coordinates}.
Longitudinal motion aligns with the LV's long axis, while radial motion involves changes in distance from the LV cavity center to the myocardial wall. Circumferential motion is characterized by circumferential shifts along the epicardial wall.

In our specific context, we determine the longitudinal direction as the principal component of the LV segmentation, intersecting the center of mass of the LV segmentation. The radial direction comprises every line normal to the principal component of the LV and intersecting with it. The circumferential component is normal to both the intersecting radial line and the main component itself.

\section{Method}

\subsection{Data Preparation}

We begin with a single cardiac 4D CT scan, containing systolic and diastolic frames, along with its associated manually segmented LV. 
We identify the principal direction of the LV within this phase using LV segmentation.

\subsection{Initial Parameters}

We systematically apply radial and longitudinal deformations along the principal direction of the LV. 
These deformations introduce variations in shape, closely mimicking physiological changes during the cardiac cycle. 
We ensure that the synthetic frame aligns with the non-synthetic counterpart in terms of LV volume, calculated based on segmentation. 
We store the parameters governing radial and longitudinal deformation.

\subsection{Torsion selection and Warping}

\begin{figure}[h] 
    \centering
        \includegraphics[width=\linewidth]{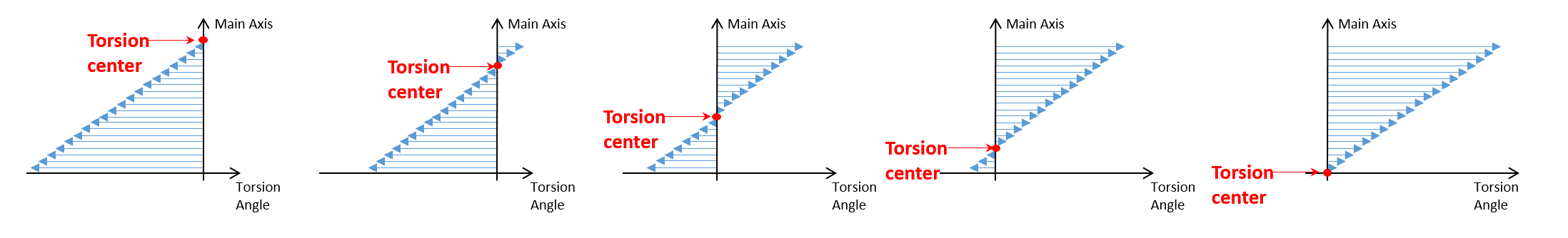}
        \caption{A visualization of five different torsion centers evenly distributed for a constant torsion.}
    \label{fig:torsion_centers}
\end{figure}

These radial and longitudinal parameters serve as the basis for generating synthetic frames, incorporating different torsions and torsion centers. 
Torsion, in this context, denotes the difference in circumferential deformation between the most positive and negative points of the LV along its main axis. The torsion center represents the point along the main axis where the torsion equals zero. Figure \ref{fig:torsion_centers} illustrates a consistent torsion with varying torsion centers.

Each set of radial, longitudinal, torsion, and torsion center values is employed to warp a single systolic frame, resulting in a pair of systolic and corresponding synthetic diastolic frames. 
The deformations are applied exclusively to pixels within the LV segmentation.

These deformations are applied to both the 4D CT scan, from which the radial and longitudinal parameters are derived, and additional 3D CT scans. Each additional scan contains a single 3D frame and a corresponding LV segmentation map.

\subsection{Storing GT annotations}

We store the 3D optical flow information representing the deformation applied during synthetic data generation. Essentially, the 3D array representing the deformation values applied to the voxels serves as the ground truth annotations for each generated image.

\section{Implementation Details}

For the initial selection of radial and longitudinal parameters, the cardiac 4D CT scan from the DICOM Image Library of \cite{osirix} called "MAGIX" was utilized. 
The LV segmentation was performed using D2P software. 
Additionally, four 3D CT scans from different patients in the Multi-Modality Whole Heart Segmentation (MM-WHS) dataset~\cite{gao2023bayeseg},~\cite{zhuang2018multivariate},~\cite{luo2022mathcal},~\cite{wu2022minimizing} were employed.
Specifically, scans numbered 1001, 1003, 1016, and 1020 from the train set were chosen for their visually apparent proper contrast. 
The longitudinal ratio of 0.9, representing the ratio of the length of the LV along the longitudinal direction between systole and diastole, was selected. 
Additionally, a radial ratio of 0.91, indicating the length in the ratio of the distance in the radial direction between systole and diastole, was chosen. 
For the selection of torsion and torsion centers, we aimed to cover a variety of common torsions observed physiologically in the LV.
Torsion centers were distributed equally along the main axis to ensure a comprehensive representation of torsion variations. 
A total of 300 synthetic frames were generated, encompassing 12 torsion values, ranging from 0 to approximately 35 degrees. Each torsion level includes five "torsion centers".

These variations in torsion and torsion centers contribute to a diverse dataset, capturing the range of physiological deformations in the LV during the cardiac cycle.

\section{Conclusion}

In this work, we presented a novel approach to synthetic data generation for enriching cardiovascular imaging datasets. The resulting dataset offers valuable GT annotations for 3D optical flow for myocardium deformation analysis. The publicly available code allows for reproducibility and further exploration in the field.

\bibliographystyle{splncs04}
\bibliography{references}

\begin{thebibliography}{1}
\providecommand{\url}[1]{\texttt{#1}}
\providecommand{\urlprefix}{URL }
\providecommand{\doi}[1]{https://doi.org/#1}

\bibitem{abduch2014cardiac}
Abduch, M.C.D., Alencar, A.M., Mathias~Jr, W., Vieira, M.L.d.C.: Cardiac mechanics evaluated by speckle tracking echocardiography. Arquivos brasileiros de cardiologia  \textbf{102},  403--412 (2014)

\bibitem{gao2023bayeseg}
Gao, S., Zhou, H., Gao, Y., Zhuang, X.: Bayeseg: Bayesian modeling for medical image segmentation with interpretable generalizability. arXiv preprint arXiv:2303.01710  (2023)

\bibitem{jeung2012myocardial}
Jeung, M.Y., Germain, P., Croisille, P., ghannudi, S.E., Roy, C., Gangi, A.: Myocardial tagging with mr imaging: overview of normal and pathologic findings. Radiographics  \textbf{32}(5),  1381--1398 (2012)

\bibitem{luo2022mathcal}
Luo, X., Zhuang, X.: An n-dimensional information-theoretic framework for groupwise registration and deep combined computing. IEEE Transactions on Pattern Analysis and Machine Intelligence  (2022)

\bibitem{osirix}
Rosset, A., Spadola, L., Ratib, O.: Osirix: An open-source software for navigating in multidimensional dicom images. Journal of digital imaging : the official journal of the Society for Computer Applications in Radiology  \textbf{17},  205--16 (10 2004). \doi{10.1007/s10278-004-1014-6}

\bibitem{wu2022minimizing}
Wu, F., Zhuang, X.: Minimizing estimated risks on unlabeled data: A new formulation for semi-supervised medical image segmentation. IEEE Transactions on Pattern Analysis and Machine Intelligence  \textbf{45}(5),  6021--6036 (2022)

\bibitem{zhuang2018multivariate}
Zhuang, X.: Multivariate mixture model for myocardial segmentation combining multi-source images. IEEE transactions on pattern analysis and machine intelligence  \textbf{41}(12),  2933--2946 (2018)

\end{thebibliography}

\end{document}